%% file: main.tex
\documentclass[letterpaper]{article} 
\usepackage{aaai23}  
\usepackage{times}  
\usepackage{helvet}  
\usepackage{courier}  
\usepackage[hyphens]{url}  
\usepackage{graphicx} 
\usepackage{natbib}  
\usepackage{caption} 
\nocopyright 

\title{Defending From Physically-Realizable Adversarial Attacks \\Through Internal Over-Activation Analysis}
\author{Giulio Rossolini, Federico Nesti, Fabio Brau, Alessandro Biondi and Giorgio Buttazzo
    \thanks{\texttt{email: name.surname@santannapisa.it}\\ Under review.}}
\affiliations{%
    Department of Excellence in Robotics and AI, Scuola Superiore Sant’Anna, Pisa, Italy\\
}

\usepackage[inline]{enumitem}
\usepackage{amsthm}
\usepackage{bbm}
\usepackage{xcolor}
\usepackage{epsfig}
\usepackage{amsmath,amssymb} 
\usepackage{cleveref}
\usepackage{subcaption} 
\usepackage{booktabs}
\usepackage{multirow}
\usepackage{tikz}
\usepackage{comment}
\usepackage{colortbl}
\usepackage{bm}

\newcommand{\LL}{\mathcal{L}}

\newcommand{\argmin}{\mathrm{arg}\!\min}
\newcommand{\x}{\textbf{x}}
\newcommand{\y}{\textbf{y}}

\newcommand{\del}[1]{\iffalse #1 \fi}
\newcommand{\add}[1]{{\color{black} #1}}

\begin{document}
\maketitle
\begin{abstract}
This work presents \textit{Z-Mask}, an effective and deterministic strategy to improve the adversarial robustness of convolutional networks against \emph{physically-realizable} adversarial attacks.
The presented defense relies on specific \textit{Z-score} analysis performed on the internal network features to detect and mask the pixels corresponding to adversarial objects in the input image. To this end, spatially contiguous activations are examined in shallow and deep layers to suggest potential adversarial regions. Such proposals are then aggregated through a multi-thresholding mechanism.
The effectiveness of \textit{Z-Mask} is evaluated with an extensive set of experiments carried out on models for semantic segmentation and object detection. The evaluation is performed with both digital patches added to the input images and printed patches in the real world.
The results confirm that \textit{Z-Mask} outperforms the state-of-the-art methods in terms of detection accuracy and overall performance of the networks under attack.
Furthermore, \textit{Z-Mask} preserves its robustness against defense-aware attacks, making it suitable for safe and secure AI applications.
\end{abstract}
\input{1_intro}
\input{2_rel}
\input{3_method}
\input{4_exp}
\input{5_conclusion}

\bibliography{main}

\end{document}

%% file: 1_intro.tex
\section{Introduction}
\label{sec:intro}

Nowadays, deep neural networks (DNNs) yield impressive performance in computer vision tasks such as semantic segmentation (SS) and object detection (OD). These remarkable results have encouraged the use of deep learning models also in \emph{cyber-physical systems} (CPS) as autonomous cars. However, the trustworthiness of neural networks is often questioned by the existence of adversarial attacks \cite{survey_trust}, especially those performed in the physical world \cite{pmlr-v80-athalye18b,DBLP:conf/eccv/WuLDG20,2022arXiv220101850R,braunegg2020apricot,kong_physgan_2021}, which are most relevant to CPS. Such attacks are usually crafted by means of adversarial objects, most often in the form of patches \cite{brown_adversarial_2018}, which are capable of corrupting the model outcome when processed as a part of the input image and can also be printed to perform \emph{physically-realizable} attacks.

To defend DNNs from these adversarial objects, several techniques were proposed in the literature based on specialized learning modules or robust training.
However, such approaches are often expensive, do not transfer well in realistic scenarios, and are still susceptible to specific attacks.

Differently from those strategies, the defense method proposed in this paper leverages the evidence that physical adversarial attacks yield anomalous activation patterns in the internal network layers.
Such anomalous activations caused by adversarial patches have been observed in several works \cite{yu2021defending,co_real,2022arXiv220101850R}, but to the best of our records, no studies deepen this phenomenon from a spatial perspective.
To this end, recalling the spatial propagation effect of CNNs \cite{NIPS2012_c399862d}, we noticed that a set of shallow layers contains high/medium over-activations in the spatial image areas corresponding to adversarial objects, while in deeper layers such over-activations grow in magnitude while referring to lower spatial resolutions (further illustrations in the supplementary material). 
Based on such evidences, this paper proposes \textit{Z-Mask}, a novel defense mechanism that combines the analysis of multiple layers to precisely detect and mask potential adversarial objects. 
\begin{figure}[tp]
\centering
\makebox[\columnwidth]{\includegraphics[width=1.1\columnwidth]{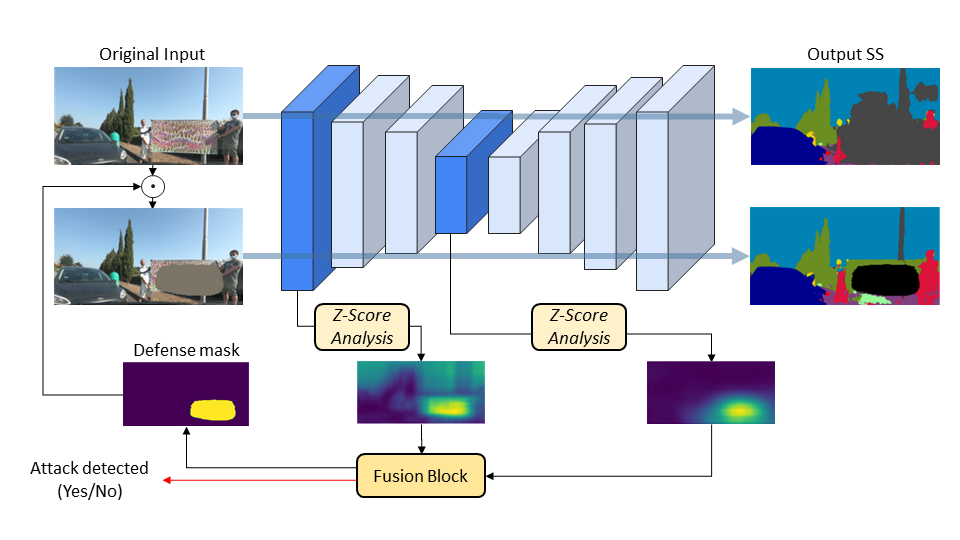}}
\caption{Illustration of the proposed approach.}
\label{fig:top_view}
\end{figure}

Figure \ref{fig:top_view} illustrates the proposed defense approach for the case of SS.
To extract preliminary adversarial region proposals, \textit{Z-Mask} runs an over-activation analysis \del{(presented in Section \ref{sec:z-heatmap})} on a set of selected layers. This analysis exploits a \textit{Spatial Pooling Refinement} (SPR) to filter out high-frequency noise in over-activated regions.
For each of these layers, the analysis produces an adversarial region proposal expressed through a heatmap. Then, all the heatmaps are aggregated into a \textit{shallow heatmap} $\mathcal{H}^{\mathcal{S}}$ and a \textit{deep heatmap} $\mathcal{H}^{\mathcal{D}}$, which summarize the over-activation behavior at two different depth levels. Finally, these two heatmaps are processed by a \textit{Fusion and Detection Block} \del{(described in the following Sections)} that flags the presence of an adversarial object and generates the corresponding defense mask.

\add{
A set of experimental results \del{(reported in Section \ref{sec:exp})}highlights the effectiveness and the robustness of the proposed defense approach on digital and real-world scenarios, which outperforms the state-of-the-art methods both in adversarial objects detection and masking. 
Furthermore, the experiments show the robustness of \textit{Z-Mask} against defense-aware attacks, which is a property inherited from the clear relation between over-activations and adversarial patches.
In summary, this paper provides the following contributions:
\begin{itemize}
    \item It proposes \textit{Z-Mask}, a novel robust adversarial defense method designed to detect and mask the regions belonging to adversarial objects; 
    \item It shows the effectiveness of a
    \textit{Z-score}-based defense by improving a naive neuron-wise approach with a \textit{Spatial Pooling Refinement}, which removes high-frequency noise and helps extract proper contiguous masks;
    \item It provides an activation-aware patch optimization to strengthen the relation between over-activations and adversarial effects induced from physical attacks.
\end{itemize}

The remainder of the paper is organized as follows: first, it introduces the related work, then it presents the \textit{Z-Mask} pipeline, it reports the experimental results, and finally states the conclusions.}

%% file: 2_rel.tex
\section{Related work}
\label{sec:rel}

\paragraph{Physical adversarial attacks.}
Adversarial attacks are widely studied methods capable of easily fooling neural models by adding imperceptible input perturbations  \cite{nakka_indirect_2020,metzen_universal_2017,DBLP:conf/iccv/XieWZZXY17,DBLP:journals/corr/SzegedyZSBEGF13,rony2019decoupling,brau2022minimal}. However, in recent years, particular interest has been devoted to adversarial attacks aimed at controlling the output of DNNs through physical adversarial objects or patches. 

In this context, Athalye et al. \shortcite{pmlr-v80-athalye18b} presented the Expectation Over Transformations (EOT) paradigm, which allows crafting adversarial objects robust against real-world transformations, as scaling, translation, orientation, and illumination changes. Later, Brown et al. \shortcite{brown_adversarial_2018} proposed an attack method based on adversarial patches, which achieved great success as a means to study the real-world robustness of DNNs and generate new effective physical attacks \cite{braunegg2020apricot,DBLP:conf/eccv/WuLDG20,9706854,lee_physical_2019,NaturalisticHu2021}.

\paragraph{Defense methods.}
To tackle the problem of physical attacks and digital adversarial patches, several defense methods have been proposed in the literature. For the sake of clarity, we divide defense methods in two main categories: \textit{adversarial training} and \textit{external tools}.
\add{The former aims at making a model more robust by re-training the network including
attacked images and regularization terms \cite{saha_role_2020,metzen2021meta,rao2020adversarial,wu_defending_2020}. These approaches significantly increase the training and testing efforts (especially when dealing with adversarial patches), without 
providing any mechanisms to notify their presence or masking their pixels. 
}

Conversely, the methods based on external tools preserve the original model parameters and complement the model output with additional information that typically consists in an attack detection flag \cite{co_real,2022arXiv220101850R,10.1145/3460120.3484757,xu_lance_2019} and/or defense masks \cite{chiang_adversarial_2021,naseer_local_2019,chou_sentinet_2020,liu2021segment,xiang2021patchguard,zhou_information_2020} that remove the adversarial parts of the image.
Although all such methods do not alter the model parameters, only a few of them are task-agnostic (e.g., capable of working for both OD and SS models) or address 
comprehensive evaluations on large datasets and realistic scenarios.

\add{
\paragraph{The role of internal activations.}
Among the large plethora of methods that study the internal behavior of DNNs under adversarial perturbations, some works~ \cite{coverage_defense,yu2021defending} noticed that adversarial patches cause large activations in the internal network layers. In particular, \cite{co_real,2022arXiv220101850R} exploited this fact to detect adversarial patches by computing the cumulative sum of the neurons activation in a certain layer. Such a score is deemed as \textit{safe} or \textit{unsafe} by comparing it to a threshold. Although this approach achieves good performance in detecting adversarial patches, it is applied to a single layer only using a neuron-wise over-activation analysis, which may leave room for effective defense-aware attacks.
Furthermore, it is limited to detection purposes only, without addressing the fact through a spatial analysis. 

\paragraph{This work }
faces the over-activation phenomenon also from a spatial perspective to derive an effective and straightforward defense that performs a multi-layer and a multi-neuron analysis. First, a \textit{spatial pooling refinement} based on the \textit{Z-Score} values is introduced in the analysis, which helps better identify the image regions that cause the over-activations. Second, shallow and deep analysis are combined to generate an aggregated defense mask.
These steps make \textit{Z-Mask} a fully task-agnostic defense that outputs both a precise pixel mask and an attack detection flag. 
It works on top of any pretrained convolutional model in the context of a large-scale evaluation that also targets realistic attacks (i.e., physically-printed patches) and preserves its robustness against defense-aware attacks.}

%% file: 3_method.tex
\section{Proposed defense}
\label{sec:method}

This section presents the \textit{Z-Mask} defense strategy, which is formulated to be task agnostic, i.e., applicable on any convolutional model. In this work, we consider the case of SS and OD models.
In both cases, the input consists of an image with $H\times W$ pixels and $C$ channels, denoted by $\x \in [0,1]^{C \times H \times W}$, while the form of the output $f(\x)$ depends on the task.
For a semantic segmentation model with $N$ classes, the output $f(\x)\in[0,1]^{N \times H\times W}$ is an image that encodes the semantic context of each pixel.
For an OD model, the output $f(\x)$ is a tensor encoding the class and the bounding box of each detected object. 
Without loss of generality, a task-specific loss function $\LL(f(\x), \y)$ is used to quantify the quality of a prediction $f(\x)$ against the ground-truth output $\y$. 

A real-world adversarial attack can be simulated by applying an \textit{adversarial patch} in a specific region of the input image $\x$. 
A patch $\bm\delta$ is a $\tilde H\times \tilde W$ image within $C$ channels, where $\tilde H \leq H$ and $\tilde W \leq W$.
Crafting an adversarial patch requires solving an optimization problem that aims at minimizing a specific attack loss function while making patch features more robust against real-world transformations in the input image \cite{pmlr-v80-athalye18b}.

In detail, given an input image $\x$ and a patch $\bm\delta$, an additional function $\gamma$ is randomly sampled from a set $\bm\Gamma$ of compositions of appearance-changing and placement transformations. The appearance-changing transformations include brightness, contrast change and noise addition; the patch placement transformations include random translation and scaling for defining the position of the patch in the image. Then, a patch $\bm\delta$ is applied to $\x$, according to $\gamma$, through a patch application function $g_{\gamma}(\x,\bm\delta)$. 
Formally, an adversarial patch $\bm{\hat\delta}$ can be crafted by solving the following optimization:
\begin{equation}
    \bm{\hat\delta} = \argmin_{\bm\delta}~\mathbb{E}_{\x\sim\mathbf{X}, \gamma\sim\mathbf{\Gamma}} ~ \mathcal{L}_{Adv}(f(g_{\gamma}(\x,\bm\delta), \y_{Adv}),
\end{equation}
where $\mathbf{X}$ is a set of known inputs, $\y_{Adv}$ is the adversarial target, and $\mathcal{L}_{Adv}$ is the adversarial loss that specifies the objective of the attacker.
In the case of untargeted attacks, the adversarial target is the regular ground truth $\y$ and the adversarial loss function is $-\LL(f(\tilde \x), \y)$, to maximize the task-specific loss.
To enhance the physical realizability of the patches, the adversarial loss includes additional terms that are described in the supplementary material. 

A defense masking strategy obscures a portion of the input image (supposedly containing the adversarial patch) through a pixel-wise product $\odot$ with a binary mask having the same size of the image.
Formally, for each perturbed image $\tilde\x = g_{\gamma}(\x,\bm \hat\delta)$, a binary mask $M(\tilde\x)$ is computed with the intent of satisfying the following property:
\begin{equation}
\LL(f(\tilde\x\odot M(\tilde\x)),\y) \approx \LL(f(\x),\y).
\label{eq:mask_goal}
\end{equation}
Equation~\eqref{eq:mask_goal} states that the objective of a masking defense is to mitigate the effectiveness of a physical adversarial perturbation while preserving a correct behavior outside the region of the mask.
\add{
In this work, Mask $M(\tilde\x)$ is generated by leveraging multiple over-activation analysis, which are then aggregated through a \textit{Fusion Block} mechanism.
}

\subsection{Layer-wise over-activation analysis}
\label{sec:z-heatmap}

\begin{figure}[tp]
\centering
\makebox[\columnwidth]{\includegraphics[width=0.5\textwidth]{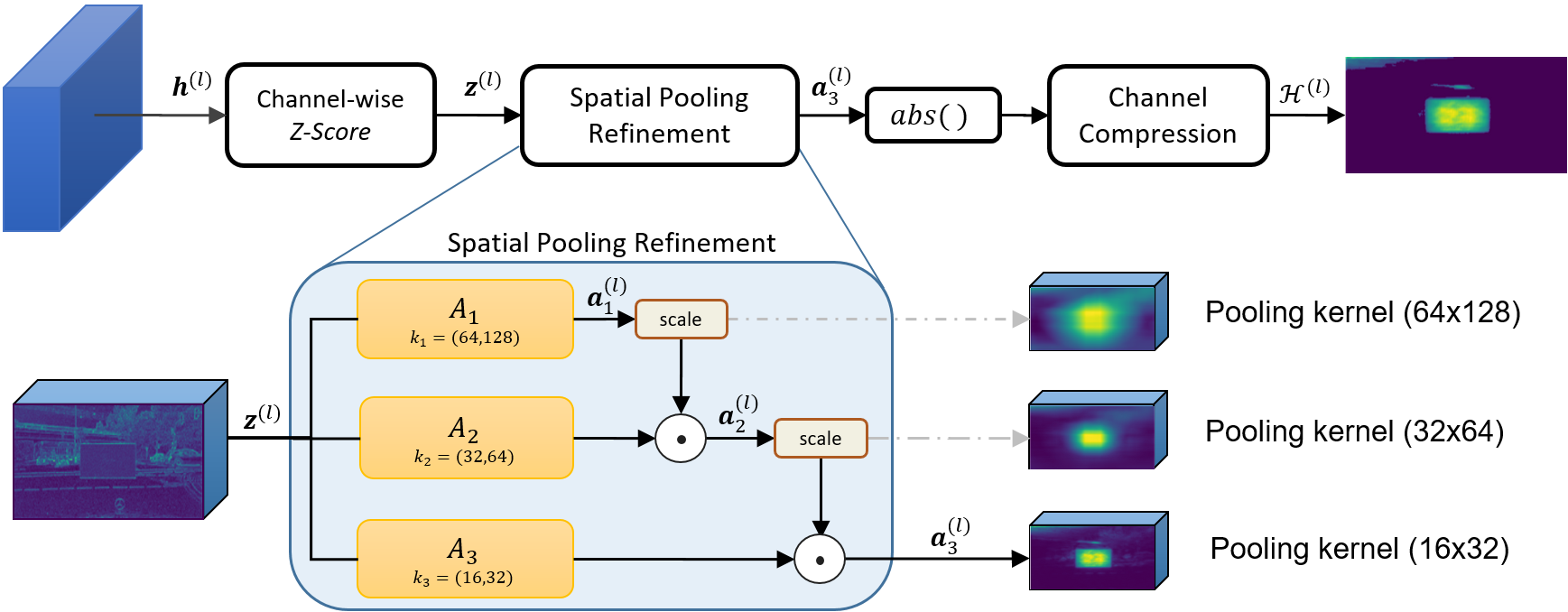}}
\caption{Over-activation pipeline performed by \textit{Z-Mask} on a given layer with $m=3$ Average-Pooling stages. The \textit{scale} blocks used in the SPR refer to the $\infty$-norm used in Equation \ref{eq:spatial-refinement}. Resizing operations are omitted in the figure. }
\label{fig:idea}
\vspace{-1.5em}
\end{figure}

Let $\bm h^{(l)} \in \mathbb{R} ^ {C^{(l)} \times H^{(l)} \times W^{(l)}} $ be the output features of layer $l$, obtained during the forward pass of $f(\x)$, where $H^{(l)}$ and $W^{(l)}$ are its spatial dimensions. The heatmap $\mathcal{H}^{(l)}$ is obtained by applying the following operations to $\bm h^{(l)}$ (illustrated in Figure~\ref{fig:idea}).
First, for a layer $l$, the channel-wise \textit{Z-score} $\bm z^{(l)} = \frac{\bm h^{(l)} - 
\mu^{(l)}}{\sigma^{(l)}}$ of $\bm h^{(l)}$ is computed, where $\mu^{(l)}$ and $\sigma^{(l)}$ are the channel-wise mean and standard deviation of the output features, respectively, obtained from a dataset $\bm X$ that does not include attacked images. The \textit{Z-score} is then processed in cascade by a sequence of $m$ Average-Pooling operations $(A_1, \ldots, A_i, \ldots A_m)$ as follows:
\begin{equation}
    \begin{cases}
        \bm a^{(l)}_i = \mathcal{R} (A_i(\mathcal{R}(\bm z^{(l)})))\odot\frac{\bm a^{(l)}_{i-1}}{\|\bm a^{(l)}_{i-1}\|_\infty} , & i=1,\ldots,m\\
        \bm a^{(l)}_{0} \equiv 1,
    \end{cases}
    \label{eq:spatial-refinement}
\end{equation}
where each $A_i$ has kernel size $k_i$ and $\mathcal{R}$ is an operator that resizes (by interpolation) the spatial dimensions of a given tensor to a configurable size $H^{\mathcal{R}} \times W^{\mathcal{R}}$.
Note that the $i^\texttt{th}$ kernel is larger than the $(i+1)^\texttt{th}$ one. Also, the resize operation is performed before and after each $A_i$ to enable the use of the pixel-wise product and the same sequence of Average-Pooling operations on different network layers\del{, which otherwise may require working with tensors of different sizes}.

The rationale for using such Average-Pooling operations is the following.
Observe that the \textit{Z-score} itself provides a pixel-wise metric capable of highlighting the over-activated pixels (i.e., pixels with 
internal activation values that are significantly far from $\mu^{(l)}$ in terms of $\sigma^{(l)}$). 
However, since we aim at masking adversarial patches, we are interested in highlighting \emph{contiguous} over-activated portions of the image rather than spurious over-activated pixels (i.e., pixels whose neighbors have activation values close to $\mu^{(l)}$). To do that, the SPR implements a cascade filtering \cite{avg-filtering} that reduces the effects of spurious over-activated pixels. 
The process is iteratively refined: first larger kernels identify macro-regions that include over-activated contiguous pixels and then smaller kernels refine the analysis within such macro-regions.
\add{
Finally, to obtain the desired heatmap $\mathcal{H}^{(l)}$ (of size $1 \times H^\mathcal{R} \times W^\mathcal{R}$), the absolute values of $\bm a^{(l)}_m$ are averaged across the channels.
As shown in the experimental section, this process yields a heatmap of the over-activated region with sharper areas (Figure \ref{fig:pooling_ablation}).}
\subsection{Fusion and Detection mechanism}
\label{sec:fusion-block}
This section explains how the mask $M(\x)$ is generated by merging the information of two sets of heatmaps, $\mathcal{S}$ and $\mathcal{D}$. The set $\mathcal{S}$ contains $N_\mathcal{S}$ heatmaps belonging to the selected shallow layers only, while $\mathcal{D}$ contains $N_\mathcal{D}$ heatmaps belonging to deeper layers and possibly to shallow layers.
Leveraging these sets of heatmaps, we reduce the analysis to two aggregated heatmaps $\mathcal{H}^\mathcal{S} = \mathcal{F}(\mathcal{S})$ and $\mathcal{H}^\mathcal{D} = \mathcal{F}(\mathcal{D})$, where $\mathcal{F(\cdot)}$ is an operator that merges multiple heatmaps belonging to a given set. In practice, a pixel-wise \textit{max} function is used for $\mathcal{F}(\cdot)$.
$\mathcal{H}^\mathcal{S}$ and $\mathcal{H}^\mathcal{D}$ summarize the over-activation behavior at different depths in the model: $\mathcal{H}^\mathcal{S}$ represents the over-activated regions in the shallow layers, while $\mathcal{H}^\mathcal{D}$ takes into consideration also deep layers.
The reason for using these two heatmaps emerged after a series of experimental observations.
From a practical perspective, $\mathcal{H}^{\mathcal{S}}$ allows highlighting the over-activated portions of the image (i.e., the regions that may contain adversarial objects): it provides a high spatial accuracy, but a limited capability of discriminating adversarial and non-adversarial regions. Conversely, $\mathcal{H}^{\mathcal{D}}$ provides a high accuracy in identifying adversarial over-activations, but with a much lower spatial accuracy. In fact,
experiments showed that over-activations coming from non-adversarial regions do not propagate their effect to deeper layers (a more detailed analysis of this effect is provided in the supplementary material). Hence, $\mathcal{H}^{\mathcal{D}}$ can be used to filter out the regions highlighted by $\mathcal{H}^{\mathcal{S}}$ that are not adversarial, yielding a more accurate heatmap.

\begin{figure}[tp]
\centering
\makebox[\columnwidth]{\includegraphics[width=\columnwidth]{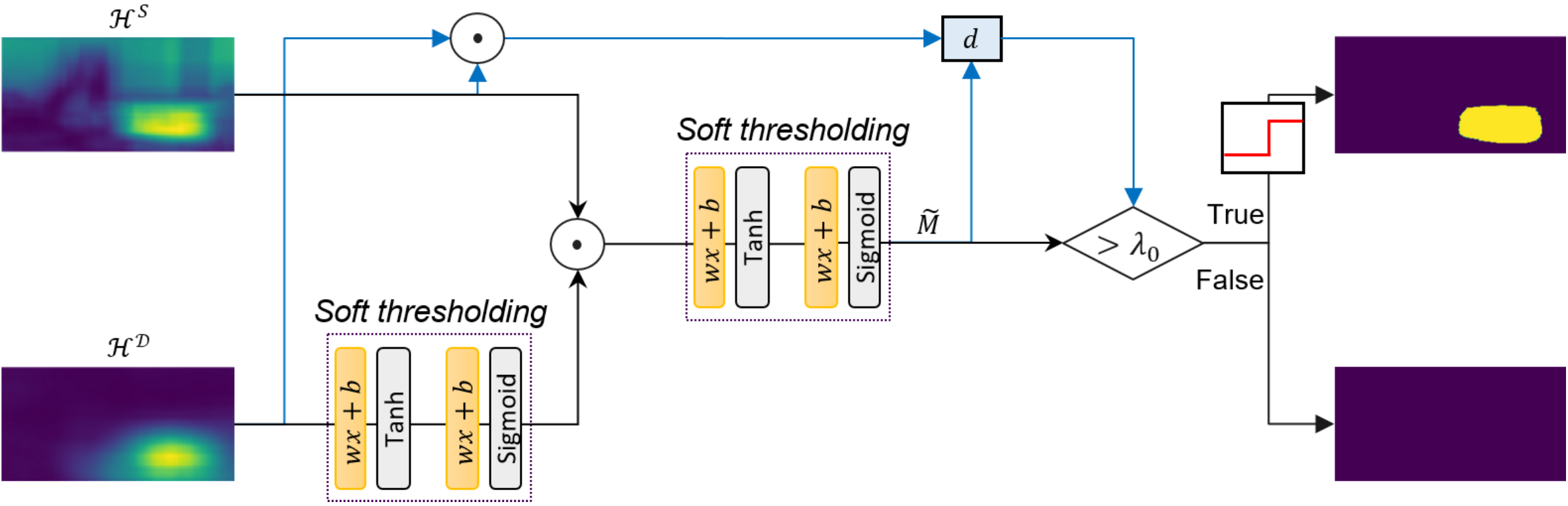}}
\caption{Fusion and Detection Block.}
\label{fig:fusion-block}
\vspace{-1.5em}
\end{figure}
Figure~\ref{fig:fusion-block} illustrates the operations performed by the Fusion and Detection Block.
The merging process leverages two \textit{soft-thresholding blocks}. The first block extracts a region of interest from $\mathcal{H}^\mathcal{D}$, which is then multiplied by $\mathcal{H}^\mathcal{S}$ to pose attention only to over-activated areas in the deeper layers. The second block extracts $\tilde{M}$, a soft version of the final mask with real pixel values in $[0,1]$. Each \textit{soft-thresholding block} consists of two sequential linear layers (both with 1-dimensional weight and bias), activated by a \textit{tanh} and a \textit{sigmoid} function, respectively.


Finally, to apply the masking only when an adversarial region is detected, we measure the over-activation as $d= \frac{\|\mathcal{H}^{\mathcal{S}}\odot \mathcal{H}^{\mathcal{D}}\odot \tilde M\|_1}{\| \tilde M\|_1}$ and compute the mask $M(\x)$ as follows:
\begin{equation}
    M(\x) =
\begin{cases} 
    1-\hslash(\tilde M),  & d>\lambda_0\\
    1,&\mbox{otherwise},
\end{cases}
\end{equation}
where $\hslash$ is the Heaviside function centered in $0.5$ and $\lambda_0$ is a given threshold. 
The soft-thresholding parameters (eight in total) are fitted by supervised learning, while the threshold $\lambda_0$ is configured through an ROC analysis. 

\add{The main motivation of using this module is its high deterministic behavior, since it mimics a soft-threshold operation in a differentiable manner. This allows testing against gradient-based defense-aware attacks and offers a transparent robustness by constraining an attacker to reduce the over-activation values to fool the defense (see experimental part).}

%% file: 4_exp.tex
\section{Experimental evaluation}
\label{sec:exp}

This section presents a set of experiments carried out on several convolutional models for OD and SS to evaluate the effectiveness of the proposed defense. 
All the experiments were implemented using PyTorch~\cite{pytorch} on a server with 8 NVIDIA-A100 GPUs.  
\add{
For both SS and OD tasks, the effectiveness of an adversarial attack was measured by evaluating the drop of the model performance with a task-dependent metric. For SS models, the \textit{mean Intersection-over-Union} (mIoU) was used on the subset of pixels not belonging to the applied patch, as done by \cite{2022arXiv220101850R}.
}
For OD models, the performance was measured by the \textit{COCO mean Average Precision} (mAP) 

\paragraph{Datasets and Models.}
Three state-of-the-art models were selected for the SS task: ICNet \cite{icnet_paper}, DDRNet \cite{ddrnet_paper}, and BiseNet \cite{bisenet_paper}, using pretrained weights provided by their authors.
For the OD task, SSD \cite{liu2016ssd}, RetinaNet \cite{lin2017focal}, and Faster R-CNN \cite{7485869} were selected from the PyTorch model zoo.  More details are in the supplementary material.

Several datasets were used for the experiments.
The Cityscapes dataset \cite{DBLP:conf/cvpr/CordtsORREBFRS16} is a canonical dataset of driving images for SS. It contains 2975 and 500 $1024\times2048$ images for training and validation, respectively.
For OD, we considered the COCO 2017 dataset~\cite{lin2014microsoft}, containing 112k and 5k images for training and validation, respectively. Being COCO a dataset of common images, pictures have different sizes, hence a network-specific resizing is required.
To assess the proposed approach on real-world scenarios, we considered APRICOT \cite{braunegg2020apricot}, which is a COCO-like dataset including more than 1000 images, each containing a physical adversarial patch for one between Faster R-CNN, RetinaNet, and SSD.

\paragraph{Attack and defense strategies.}
Different attack methodologies were used to craft adversarial patches. For SS models, we leveraged the untargeted attack pipeline used in \cite{2022arXiv220101850R}, while, for OD models, we performed an untargeted attack on the classes, similarly to \cite{shapeshifter}. 
The patches contained in the APRICOT dataset rely on a false-detection attack \cite{braunegg2020apricot}.
More details are provided in the supplementary material.

Concerning defense strategies, we compared \textit{Z-Mask} against different approaches for both adversarial pixel masking and detection. For the masking task, we re-implemented the Local Gradient Smoothing method (LGS) \cite{naseer_local_2019} and MaskNet \cite{chiang_adversarial_2021}, both with the original settings described by the authors. For the adversarial detection, we considered for comparisons FPDA \cite{2022arXiv220101850R} and HN \cite{co_real}. Details are in the supplementary material.

\begin{figure}[tp]
{\includegraphics[width=\columnwidth]{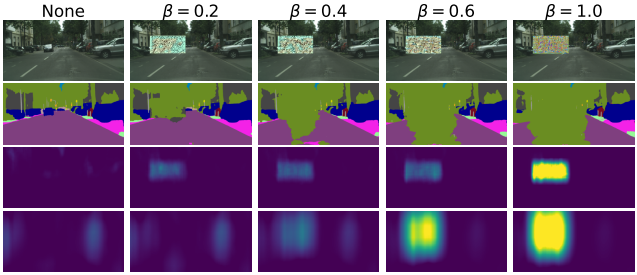}}
\caption{Visualization of the predictions and SPR heatmaps obtained from several $\beta$ activation-aware patches. The rows report $x$, $f(x)$, $\mathcal{H}^\mathcal{S}$ and $\mathcal{H}^\mathcal{D}$, respectively.}
\label{fig:illustration_beta}
\end{figure}

\paragraph{Activation-aware patch optimization.}
\add{
We crafted adversarial patches while controlling the over-activation to better understand its relation with the induced adversarial effect, as well train the defense method to properly scale into real-world scenarios. 
To do that, we proposed the following optimization: 
\begin{equation}
\begin{aligned}
\small
    \bm{\hat\delta}_{\beta} = \argmin_{\bm\delta\quad} \mathbb{E}_{\x\sim\mathbf{X}, \gamma\sim\mathbf{\Gamma}} 
    \big[
    & (1-\beta)\cdot\LL_{\textit{OZ}}(f, g_{\gamma}(\x, \bm\delta))\\
    &+\beta\cdot\LL_{\textit{Adv}}(f(g_{\gamma}(\x, \bm\delta)), \y)
    \big],
\end{aligned}
\label{eq:train_LOV}
\end{equation} 
\noindent where $\beta \in [0, 1]$ is a control parameter and $\LL_{\textit{OZ}}$ is a loss function that measures the magnitude of over-activation of internal layers
(details are available in the supplementary material).
The rationale behind this optimization problem is that a low value of $\beta$ reduces the importance assigned to the adversarial effect, while forcing the attack to generate less over-activation in the internal layers, hence simulating real-world patches.
Figure \ref{fig:illustration_beta} illustrates the over-activation of these patches (computed with the SPR) both in shallow and deep layers, remarking the relation with the induced adversarial effect. Furthermore, Figure \ref{fig:test_detection} (discussed later) provides a measure of the adversarial effect as a function of $\beta$.}

\begin{figure}[ht]
     \centering
     \begin{subfigure}{\columnwidth}
        \centering
        \includegraphics[width=\textwidth]{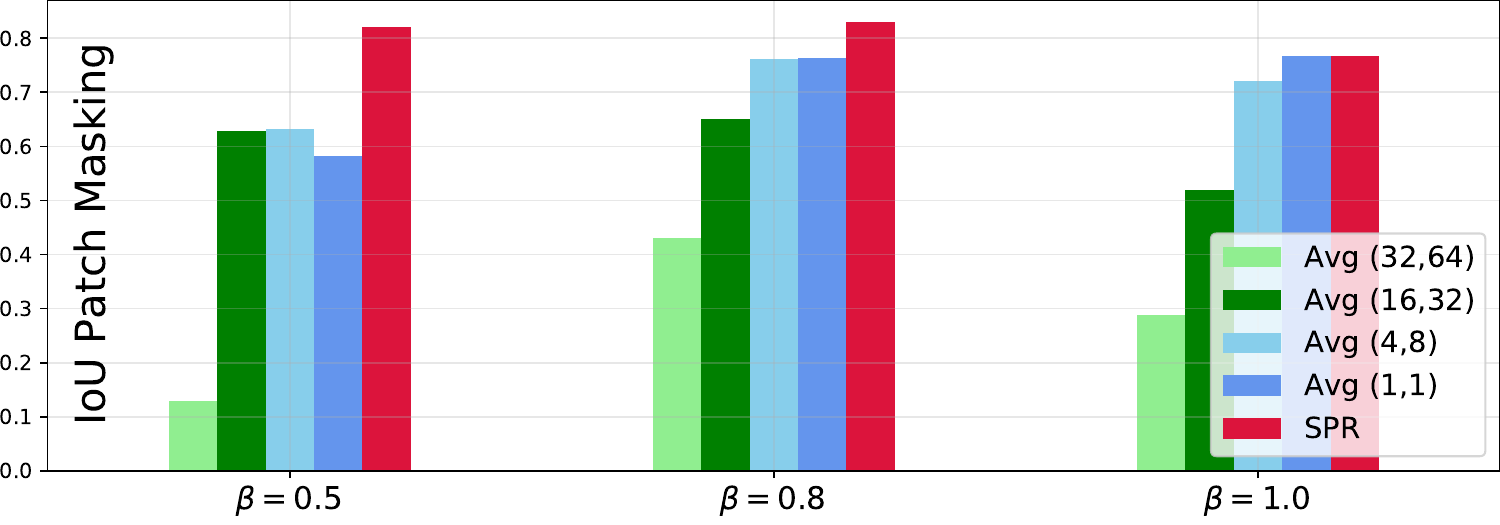}
        \caption{IoU Patch Masking comparison}
     \end{subfigure}
     \centering
     \begin{subfigure}{0.32\columnwidth}
         \centering
         \includegraphics[width=\textwidth]{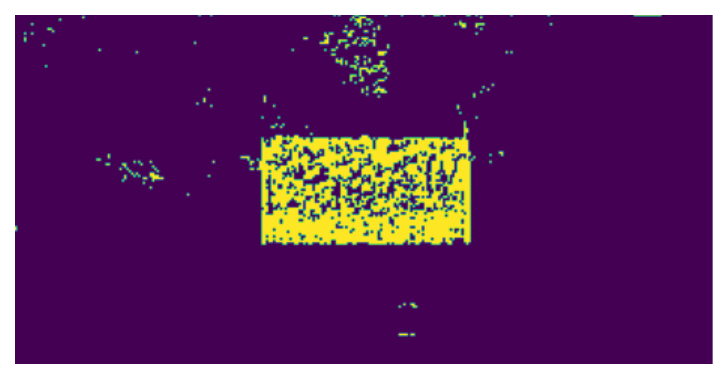}
         \caption{No Avg (1,1)}
     \end{subfigure}
     \begin{subfigure}{0.32\columnwidth}
         \centering
         \includegraphics[width=\textwidth]{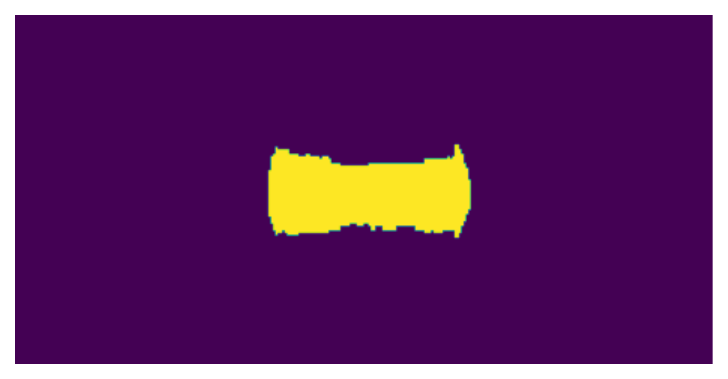}
         \caption{Avg (16,32)}
     \end{subfigure}
     \begin{subfigure}{0.32\columnwidth}
         \centering
         \includegraphics[width=\textwidth]{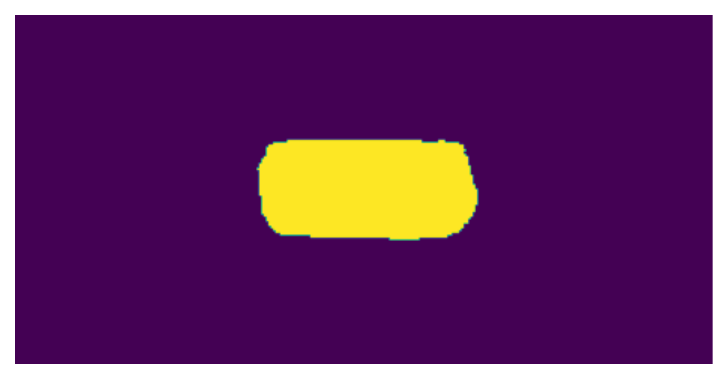}
         \caption{SPR}
     \end{subfigure}
        \caption{
        Ablation studies on the masking accuracy with different pooling strategies on 100 images of the Cityscapes validation set and BiseNet (a). Bottom figures are the predicted masks of a same input, using a patch with $\beta = 0.5$.
        }
    \label{fig:pooling_ablation}
\end{figure}

\paragraph{\textit{Z-Mask} settings and training.}
For SS models, the heatmaps in $\mathcal{S}$ were generated with a SPR composed of four pooling operations, with kernel sizes $k_1=(64, 128),\,k_2=(32, 64),\,k_3=(16, 32),\,k_4=(8, 16)$. Instead, the heatmaps in $\mathcal{D}$ were generated using two pooling operations with kernel sizes $k_1=(64, 128),\,k_2=(32, 64).$
After each pooling operation, the heatmaps were resized to $(H^\mathcal{R} \times W^{\mathcal{R}}) = (150\times300)$. Please note that all the resulting heatmaps have a 1:2 aspect ratio, keeping the same aspect ratio of the input images.
For OD models, the SPR used $k_1=(40, 40),\,k_2=(25, 25),\,k_3=(10, 10)$ to build $\mathcal{S}$, and $k_1=(80, 80),\,k_2=(40, 40)$ to build $\mathcal{D}$. The resizing dimension was set to $(400\times500)$. 
For all the tests, pooling operations were applied with stride 1.
These kernel settings were motivated by extensive preliminary tests performed to analyse the internal activations. 
\add{
To illustrate the benefits of the SPR, Figure \ref{fig:pooling_ablation} provides the results of ablation studies by comparing the performance of the Fusion and Detection block with different pooling settings and patches crafted with different $\beta$. The SPR block always achieves a better IoU Patch Masking, which is computed as the IoU between the predicted mask and its ground truth.}

The description of the layers selected for extracting $\mathcal{D}$ and $\mathcal{S}$ in each model is reported in the supplementary material.

The parameters of the \textit{soft-thresholding} operations inside the Fusion and Detection block were trained in a supervised fashion by considering a set of patches crafted with Equation \ref{eq:train_LOV} 
and minimizing the pixel-wise binary cross-entropy loss
$\mathcal{L}_{\text{BCE}}(\tilde M,\bar M)$, where $\bar M$ is the ground-truth binary mask. 
%
\add{To this end, we collected a set of adversarial patches $\Delta = \{\bm{\hat\delta}_{\beta}\,:\, \beta \in [\beta_0, 1]\}$, where we set $\beta_0=0.5$
to avoid generating patches with scarce adversarial effect.
These patches were used to craft the set $\bm{{\tilde X}}$, which was obtained by adding the patches in $\Delta$ to each image of $\bm{{X}}$. Set $\bm{{\tilde X}}$ was used to train the Fusion and Detection Block and make it robust to a wide spectrum of over-activations.}

In our tests, $\bm{{X}}$ contained $500$ images randomly sampled from the original training dataset. 
The ADAM optimizer \cite{adam} was used for this purpose, with a learning-rate of $0.01$ and training for $15$ epochs. 
The channel-wise std and mean of each selected layer was computed on a different subset of the training set containing 500 clean (i.e., non-patched) images. 
The detection threshold $\lambda_0$ was deduced after the soft-thresholding training as the \textit{cut-off} threshold of the ROC curve. The ROC was generated by computing the measure $d$ on each input of a dataset, including the clean set $\bm{{X}}$ and the patched set $\bm{{\tilde X}}$, labeled as negative and positive samples, respectively.


\input{table/digital_attack}

\subsection{Evaluation for digital attacks}
\paragraph{Masking performance.}
The benefits of the proposed defense mechanism were evaluated by attacking the validation sets with different adversarial patch sizes. For Cityscapes, we used patches with size 600x300 (L), 400x200 (M) and 300x150 (S) pixels, whereas for COCO, due to the different image aspect ratio, we used 200x200 (L), 150x150 (M), and 100x100 (S). Also, an L-size random patch was evaluated to test the case in which a portion of the image is occluded without the intent of generating an adversarial attack.

\begin{figure}[ht]
     \centering
     \begin{subfigure}{\columnwidth}
     \begin{subfigure}{1.0\textwidth}
     \begin{subfigure}{0.49\textwidth}
         \centering
         \includegraphics[width=\textwidth]{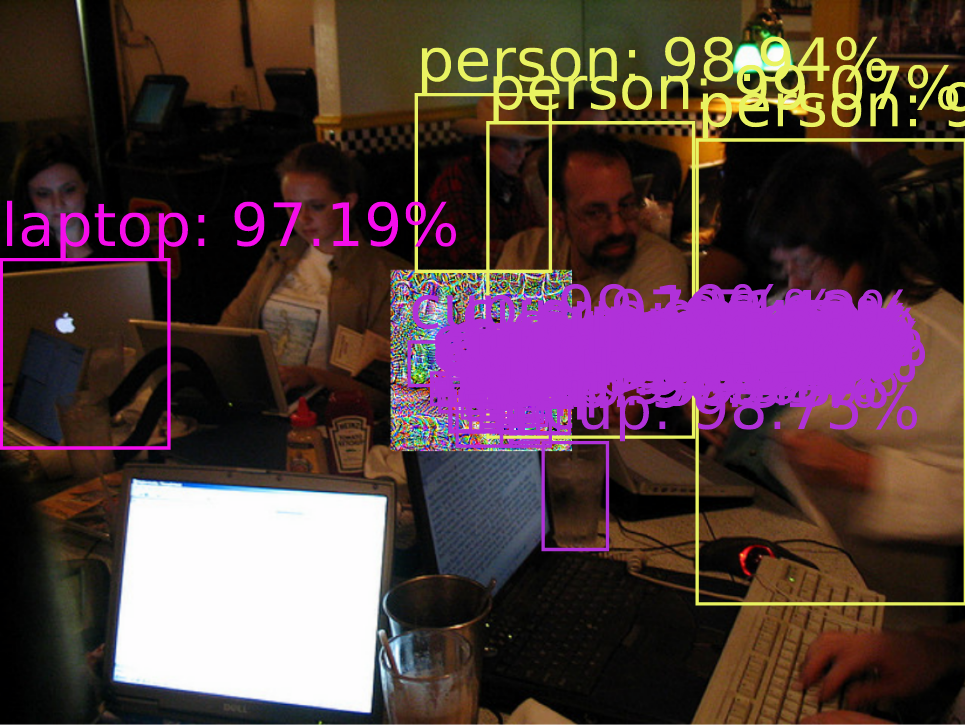}
     \end{subfigure}
     \begin{subfigure}{0.49\textwidth}
         \centering
         \includegraphics[width=\textwidth]{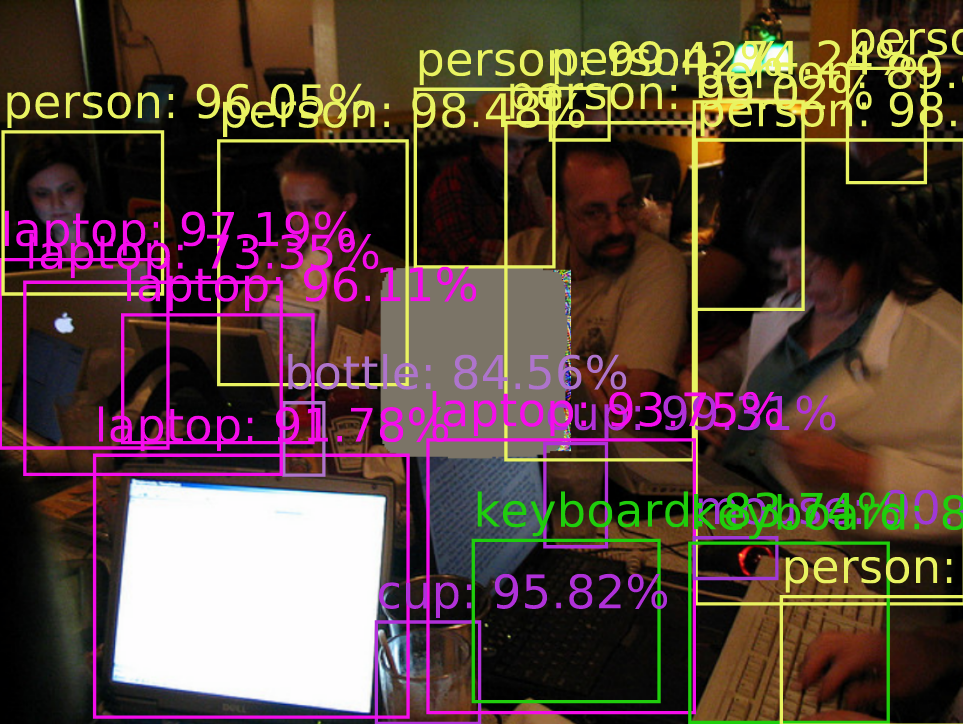}
     \end{subfigure}
     \end{subfigure}
     \caption{Faster R-CNN - COCO dataset}
     \end{subfigure}
     \begin{subfigure}{\columnwidth}
     \begin{subfigure}{0.495\textwidth}
         \centering
         \includegraphics[width=\textwidth]{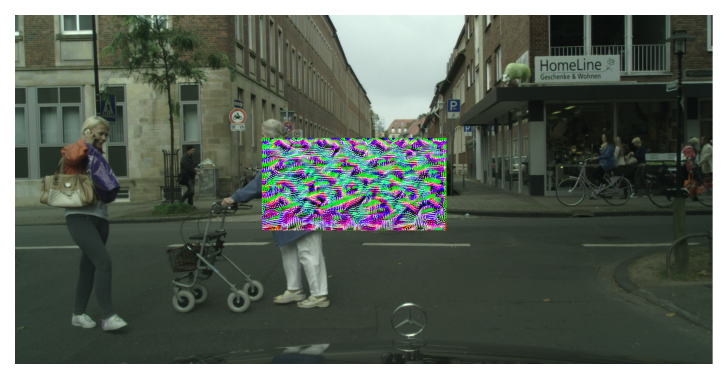}
     \end{subfigure}
     \begin{subfigure}{0.495\textwidth}
         \centering
         \includegraphics[width=\textwidth]{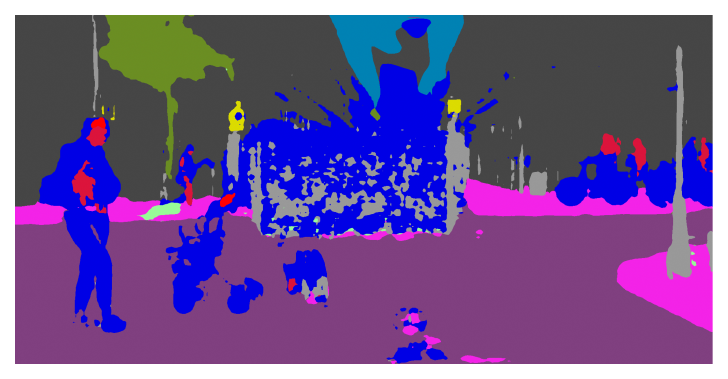}
     \end{subfigure}
     \begin{subfigure}{0.495\textwidth}
         \centering
         \includegraphics[width=\textwidth]{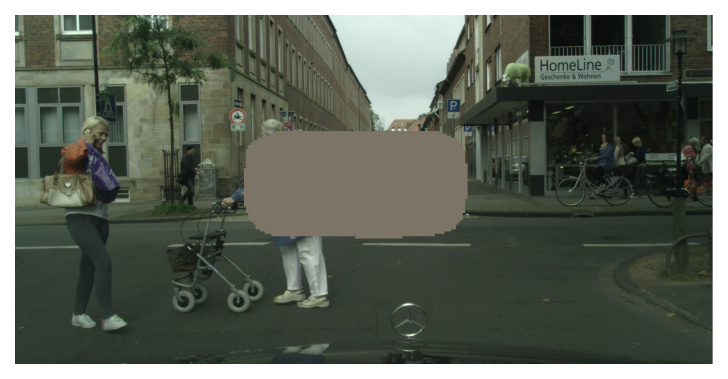}
     \end{subfigure}
     \begin{subfigure}{0.495\textwidth}
         \centering
         \includegraphics[width=\textwidth]{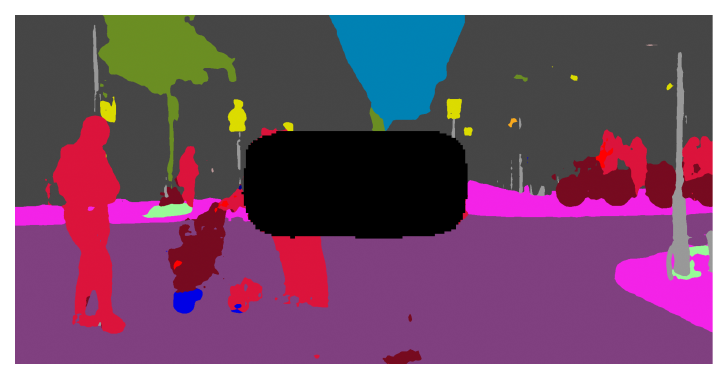}
     \end{subfigure}
     \caption{BiseNet - Cityscapes dataset}
     \end{subfigure}
     \caption{\textit{Z-Mask} effects (comparison w/ and w/o defense)}
    \label{fig:digital_images}
\end{figure}

\add{
As shown in Table \ref{table:defense_results}, \textit{Z-Mask} outperformed the other defense strategies, achieving scores similar to the random case, when tested against adversarial attacks, and close to the original model without applying patches, meaning that does not affect the nominal model performance.
Figure \ref{fig:digital_images} illustrates the benefits of \textit{Z-Mask} on images attacked digitally. Attacked areas are identified and covered without affecting other portions.
}

\begin{figure}[ht]
     \centering
     \begin{subfigure}{\columnwidth}
         \centering
         \includegraphics[width=\textwidth]{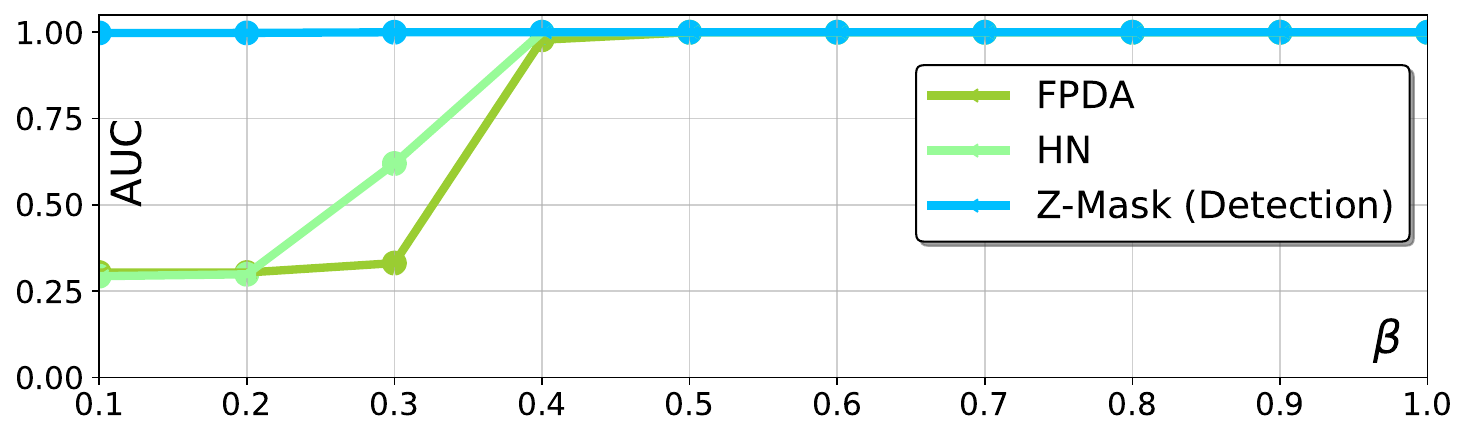}
     \end{subfigure}
     \begin{subfigure}{\columnwidth}
         \centering
         \includegraphics[width=\textwidth]{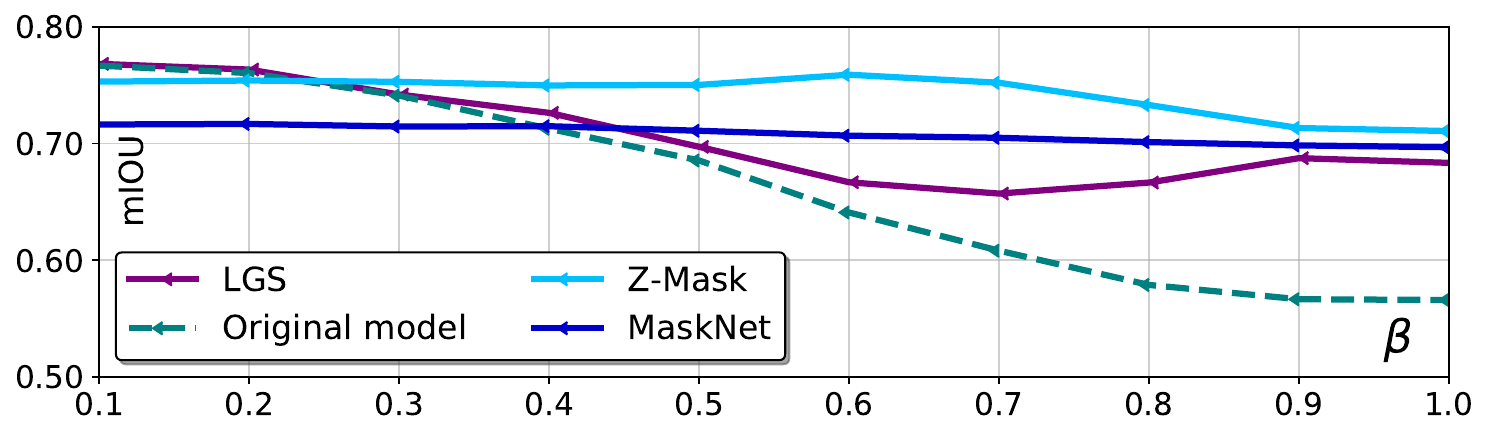}
     \end{subfigure}
        \caption{Comparison of the Detection accuracy (top plot) and task mIoU performance (bottom plot) using DDRNet on the validation set of Cityscapes.}
        \label{fig:test_detection}
\end{figure}
\paragraph{Detection performance.} All the adversarial patches evaluated in Table \ref{table:defense_results} were perfectly detected by both \textit{Z-Mask}, HN, and FPDA.
To better assess the performance of these adversarial detection methods, we used the optimization described in Equation \eqref{eq:train_LOV} to generate a set of patches with a wider range of over-activation values, selecting the values of $\beta \in \{0.1, 0.2,...,0.9,1.0\}$. Please note that $\beta=1.0$ corresponds to a regular adversarial attack, while lower $\beta$ values decrease the importance of adversarial effect to reduce the magnitude of over-activation.
An L-sized patch was generated for each $\beta$.
Figure \ref{fig:test_detection} shows the detection and masking accuracy against this set of patches as a function of $\beta$ for DDRNet. The top part of the figure shows the detection accuracy, evaluated using the AUC of ROC on a dataset, including both the clean and the attacked validation set (as negative and positive samples, respectively). Note that \textit{Z-Mask} achieved better results than the other adversarial patch detectors, providing good detection performance also to patches that do not retain much adversarial effect. 

The bottom part of the Figure \ref{fig:test_detection} reports the performance of \textit{Z-Mask}, MaskNet, LGS, and the original model (without defense). Again, our method achieved higher mIoU among almost all the $\beta$ values. 
\add{
Similar results were obtained for other models in the supplementary material.
}

\subsection{Evaluation for physical attacks}
The masking and detection performance of \textit{Z-mask} was evaluated in real-world scenarios with images containing printed adversarial patches. For this test, we adopted the same \textit{Z-mask} settings and parameters used for digital attacks on COCO, which generalize well also for real-world patches.
The detection performance was assessed with the APRICOT dataset, as positive samples, and 1000 images of the COCO validation set, as negative samples. Figure \ref{fig:ROC_apricot} (a) reports the corresponding ROC, where \textit{Z-Mask} obtained the best AUC with respect to FPDA and HN on both RetinaNet and Faster R-CNN.
The analysis on SSD was omitted since the large rescaling factor on the input image required by the pretrained network restrained APRICOT patches to just a few pixels, thus neutralizing their adversarial effect.

\add{Figure \ref{fig:ROC_apricot} (b) illustrates the effect of \textit{Z-Mask} on a sample of APRICOT. 
We also provide additional illustrations of real-world attacked datasets in the supplementary material.}
\input{table/table_apricot}

\subsection{Defense-aware attacks}
Since the \textit{Z-Mask} pipeline is fully differentiable up to $\tilde M$, an attacker might exploit that knowledge to craft defense-aware attacks, i.e., optimize patches that are adversarial for the model and the defense together. 
To this end, we propose two different defense-aware attacks.

The first attack, denoted as \textit{Mask-Attack}, is designed to induce errors in the mask output to yield an incorrect input masking operation. This would allow the adversarial patch to pass without being masked or induce additional occlusion in the image. This attack is obtained by solving the following problem with $\alpha \in \{0, 0.1, 0.2, \ldots, 1.0\}$:
\begin{equation}
\small
\begin{aligned}
    \bm{\hat\delta}_{\alpha} = \argmin_{\bm\delta\quad} \mathbb{E}_{\x\sim\mathbf{X}, \gamma\sim\mathbf{\Gamma}} \big[
    &(1-\alpha) \cdot(-\LL_{\textit{BCE}}(\tilde M, \bar M))\\
    &+ \alpha \cdot  \LL_{\textit{Adv}}(f(g_{\gamma}(\x, \bm\delta)), \y)
    \big].
\end{aligned}
\end{equation} 
Recall that $\LL_{\text{BCE}}(\tilde M,\bar M)$  is the pixel-wise binary cross-entropy loss between the defense mask $\tilde M$ and the ground-truth patch mask $\bar M$ (which is known).

\begin{figure}[tp]
     \centering
     \begin{subfigure}{\columnwidth}
         \centering
         \includegraphics[width=\textwidth]{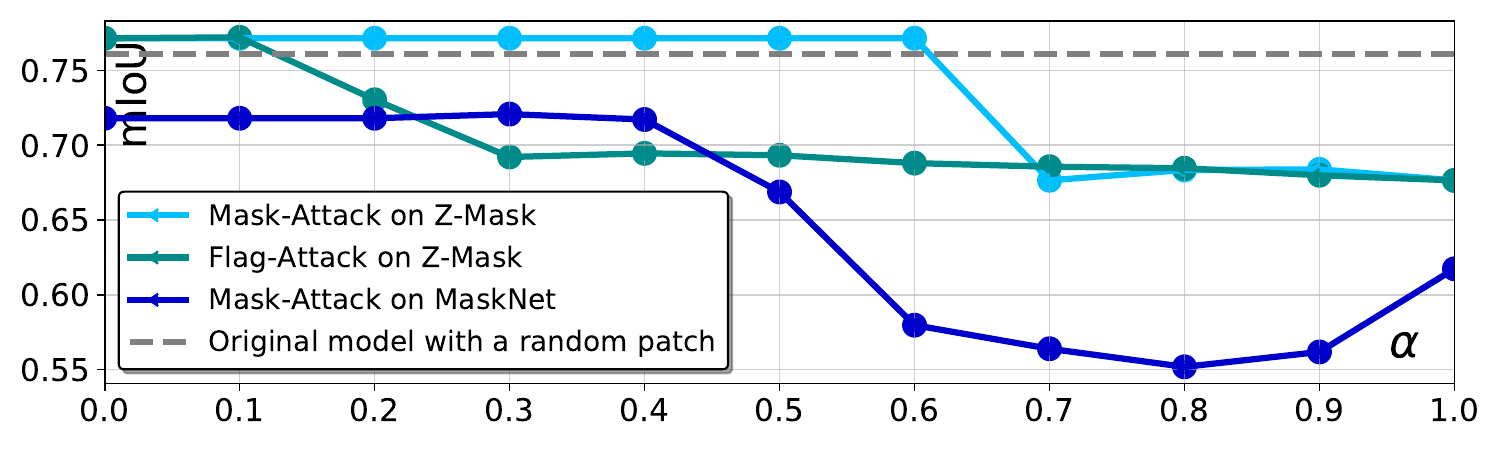}
     \end{subfigure}
     \begin{subfigure}{\columnwidth}
         \centering
         \includegraphics[width=\textwidth]{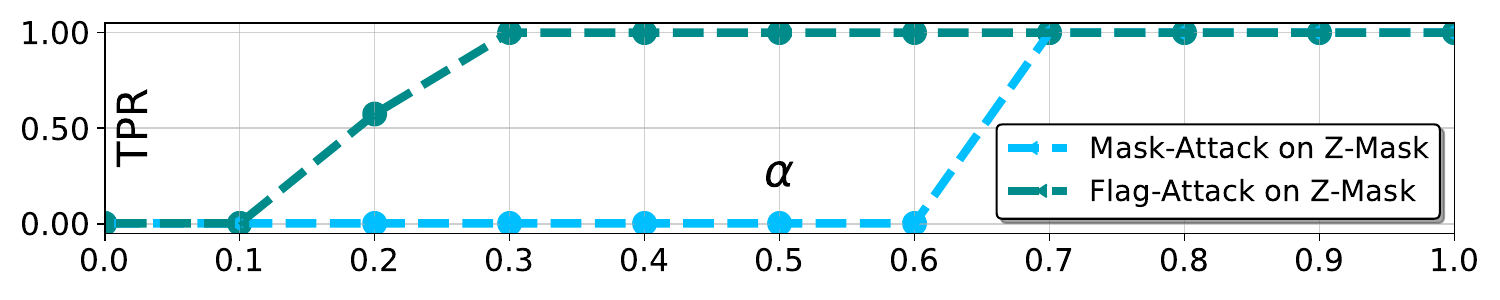}
     \end{subfigure}
        \caption{\add{Evaluation and comparison of defense benefits (mIoU) and detection performance (TPR) against defense-aware attacks as a function of $\alpha$. 
         The results refer to DDRNet evaluated on the validation set of Cityscapes.}}
        \label{fig:detection_defaware}
\end{figure}

A second attack formulation, denoted as \textit{Flag-Attack}, targets the detection flag aiming at causing false negatives in the detector. This attack is performed by replacing $\LL_{\text{BCE}}(\tilde M,\bar M)$ with
$\LL_{\text{BCE}}(\textit{Sigmoid}(d-\lambda_0), 1)$. This is done to force $d<\lambda_0$ in the optimization, hence resulting in a mask $M(\x) = 1$.
\add{
Figure \ref{fig:detection_defaware} shows the results of \textit{Z-Mask} against these attacks on DDRNet (results on other networks in the supplementary material). A mask defense-aware attack was also tested on \textit{MaskNet} to provide a comparison, while the results of LGS are not reported, since other works already addressed its weaknesses under defense-aware attacks \cite{chiang_adversarial_2021}.
%

Note that, even exploiting the knowledge of the defense, the proposed attacks were not able to reduce the performance of \textit{Z-Mask} more than what obtained for the digital evaluation, as reported in Table \ref{table:defense_results}. Indeed, observe from Figure \ref{fig:detection_defaware} that, when \textit{Z-Mask} does not detect the attack (TPR=0), the attack is not effective (maximum mIoU). 
Practically speaking, the robustness of \textit{Z-Mask} comes from the fact that it directly exploits the over-activation values.
In fact, recalling that physical attacks are strictly related to over-activations, the attacker is
required to reduce their magnitude to bypass the defense, thus inevitably yielding less effective attacks.
Conversely, for MaskNet, certain values of $\alpha$ induce larger performance degradation. 
}



%% file: table/digital_attack.tex
\begin{table}[ht]
    \footnotesize
     \begin{subtable}{\columnwidth}
     \centering
         \begin{tabular}{|c|c|cccc|}
            \hline
            Net                       & Patch & \multicolumn{4}{c|}{\begin{tabular}[c]{@{}c@{}}Defense Method (mAP Val)\end{tabular}}                       \\
                                        &               & \multicolumn{1}{c|}{\textbf{Z-Mask}} & \multicolumn{1}{c|}{MaskNet}     & \multicolumn{1}{c|}{LGS}         & None        \\ \hline
            \multirow{5}{*}{\rotatebox[origin=c]{90}{FR-CNN}}& None           & \multicolumn{1}{c|}{\textbf{0.357}}     & \multicolumn{1}{c|}{0.353} & \multicolumn{1}{c|}{0.350}  & {0.357} \\ 
                                        & Rand           & \multicolumn{1}{c|}{0.301}     & \multicolumn{1}{c|}{0.295} & \multicolumn{1}{c|}{\textbf{0.320}}  & {0.308} \\ 
                                        & S              & \multicolumn{1}{c|}{0.335}     & \multicolumn{1}{c|}{0.333} & \multicolumn{1}{c|}{\textbf{0.354}}  & {0.337} \\ \
                                        & M              & \multicolumn{1}{c|}{\textbf{0.302}}     & \multicolumn{1}{c|}{0.289} & \multicolumn{1}{c|}{0.246}  & {0.140} \\ 
                                        & L              & \multicolumn{1}{c|}{\textbf{0.300}}     & \multicolumn{1}{c|}{0.289} & \multicolumn{1}{c|}{0.244}  & {0.164} \\ \hline
            \multirow{5}{*}{\rotatebox[origin=c]{90}{SSD}} & None           & \multicolumn{1}{c|}{\textbf{0.253}}     & \multicolumn{1}{c|}{0.180} & \multicolumn{1}{c|}{0.243}  & {0.264} \\  
                                        & Rand           & \multicolumn{1}{c|}{\textbf{0.208}}     & \multicolumn{1}{c|}{0.132} & \multicolumn{1}{c|}{0.198}  & {0.215} \\  
                                        & S              & \multicolumn{1}{c|}{\textbf{0.237}}     & \multicolumn{1}{c|}{0.159} & \multicolumn{1}{c|}{0.233}  & {0.245} \\  
                                        & M              & \multicolumn{1}{c|}{\textbf{0.202}}     & \multicolumn{1}{c|}{0.125} & \multicolumn{1}{c|}{0.144}  & {0.065} \\ 
                                        & L              & \multicolumn{1}{c|}{\textbf{0.205}}     & \multicolumn{1}{c|}{0.113} & \multicolumn{1}{c|}{0.163}  & {0.072} \\ \hline
            \multirow{5}{*}{\rotatebox[origin=c]{90}{RetinaNet}} & None           & \multicolumn{1}{c|}{\textbf{0.355}}     & \multicolumn{1}{c|}{0.269} & \multicolumn{1}{c|}{0.337}  & {0.359} \\  
                                        & Rand           & \multicolumn{1}{c|}{0.305}     & \multicolumn{1}{c|}{0.227} & \multicolumn{1}{c|}{\textbf{0.312}}  & {0.308} \\ 
                                        & S              & \multicolumn{1}{c|}{\textbf{0.339}}     & \multicolumn{1}{c|}{0.245} & \multicolumn{1}{c|}{0.339}  & {0.335} \\ 
                                        & M              & \multicolumn{1}{c|}{\textbf{0.326}}     & \multicolumn{1}{c|}{0.222} & \multicolumn{1}{c|}{0.306}  & {0.304} \\ 
                                        & L         & \multicolumn{1}{c|}{\textbf{0.305}}     & \multicolumn{1}{c|}{0.212} & \multicolumn{1}{c|}{0.297}  & {0.283} \\ \hline
    \end{tabular}
    \caption{}
    \end{subtable}
    \begin{subtable}{\columnwidth}
        \centering
         \begin{tabular}{|c|c|cccc|}
            \hline
            Net                       & Patch & \multicolumn{4}{c|}{\begin{tabular}[c]{@{}c@{}}Defense Method (mIoU Val)\end{tabular}}                       \\
                                        &               & \multicolumn{1}{c|}{\textbf{Z-Mask}} & \multicolumn{1}{c|}{MaskNet}     & \multicolumn{1}{c|}{LGS}         & None        \\ \hline
            \multirow{5}{*}{\rotatebox[origin=c]{90}{DDRNet}}& None           & \multicolumn{1}{c|}{\textbf{0.778}}     & \multicolumn{1}{c|}{0.739} & \multicolumn{1}{c|}{0.777}  & {0.778}  \\
                                        & Rand           & \multicolumn{1}{c|}{0.731}     & \multicolumn{1}{c|}{0.710} & \multicolumn{1}{c|}{\textbf{0.769}}  & {0.761} \\  
                                        & S              & \multicolumn{1}{c|}{\textbf{0.741}}     & \multicolumn{1}{c|}{0.701} & \multicolumn{1}{c|}{\textbf{0.741}}  & {0.702}  \\ 
                                        & M              & \multicolumn{1}{c|}{\textbf{0.723}}     & \multicolumn{1}{c|}{0.699} & \multicolumn{1}{c|}{0.719}  & {0.663}  \\ 
                                        & L              & \multicolumn{1}{c|}{\textbf{0.691}}     & \multicolumn{1}{c|}{0.689} & \multicolumn{1}{c|}{0.642}  & {0.532}  \\\hline
            \multirow{5}{*}{\rotatebox[origin=c]{90}{BiseNet}} & None           & \multicolumn{1}{c|}{0.684}     & \multicolumn{1}{c|}{0.622} & \multicolumn{1}{c|}{\textbf{0.685}}  & {0.687} \\ 
                                        & Rand           & \multicolumn{1}{c|}{0.650}     & \multicolumn{1}{c|}{0.569} & \multicolumn{1}{c|}{\textbf{0.668}}  & {0.653}   \\  
                                        & S              & \multicolumn{1}{c|}{\textbf{0.663}}     & \multicolumn{1}{c|}{0.560} & \multicolumn{1}{c|}{0.522}  & {0.475}  \\ 
                                        & M             & \multicolumn{1}{c|}{\textbf{0.653}}     & \multicolumn{1}{c|}{0.550} & \multicolumn{1}{c|}{0.413}  & {0.323}   \\ 
                                        & L              & \multicolumn{1}{c|}{\textbf{0.621}}     & \multicolumn{1}{c|}{0.535} & \multicolumn{1}{c|}{0.320}  & {0.220}  \\ \hline
            \multirow{5}{*}{\rotatebox[origin=c]{90}{ICNet}} & None          & \multicolumn{1}{c|}{\textbf{0.785}}     & \multicolumn{1}{c|}{0.783} & \multicolumn{1}{c|}{0.782}  & {0.785} \\ 
                                        & Rand           & \multicolumn{1}{c|}{\textbf{0.768}}     & \multicolumn{1}{c|}{0.736} & \multicolumn{1}{c|}{0.764}  & {0.746}   \\ 
                                        & S              & \multicolumn{1}{c|}{\textbf{0.748}}     & \multicolumn{1}{c|}{0.737} & \multicolumn{1}{c|}{0.657}  & {0.625}  \\ 
                                        & M             & \multicolumn{1}{c|}{\textbf{0.729}}     & \multicolumn{1}{c|}{0.718} & \multicolumn{1}{c|}{0.593}  & {0.549}   \\ 
                                        & L              & \multicolumn{1}{c|}{\textbf{0.747}}     & \multicolumn{1}{c|}{0.725} & \multicolumn{1}{c|}{0.528}  & {0.430}   \\ \hline
            
        \end{tabular}
        \caption{}
    \end{subtable}
    \caption{Robustness performance evaluated for different patch sizes for OD-COCO (a) and SS-Cityscapes (b).}
    \label{table:defense_results}
\end{table}%

%% file: table/table_apricot.tex
\begin{figure}[ht]
    \begin{subfigure}{\columnwidth}
        \includegraphics[width=\columnwidth]{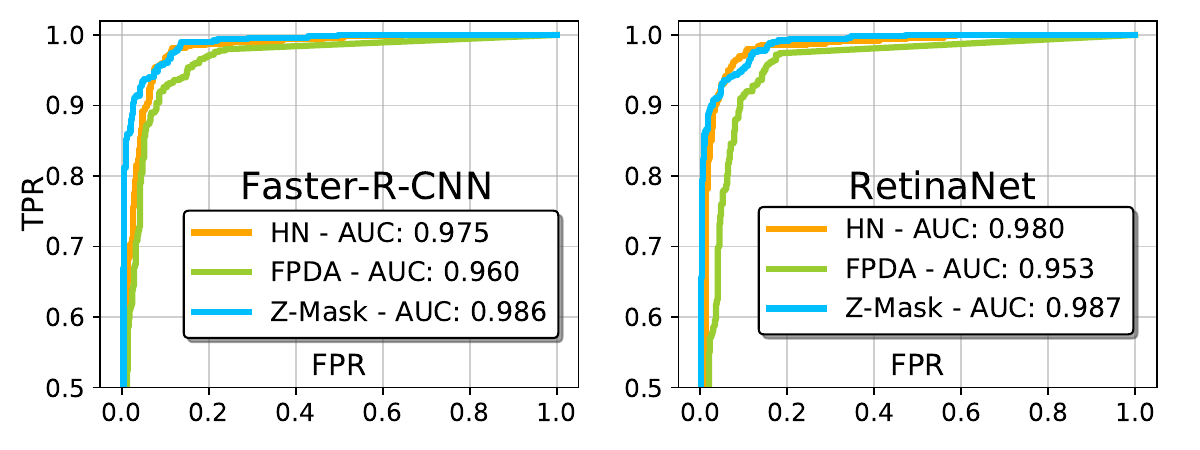}
        \caption{}
    \end{subfigure}
    \begin{subfigure}{\columnwidth}
        \begin{subfigure}{0.5\columnwidth}
            \includegraphics[width=1.\textwidth]{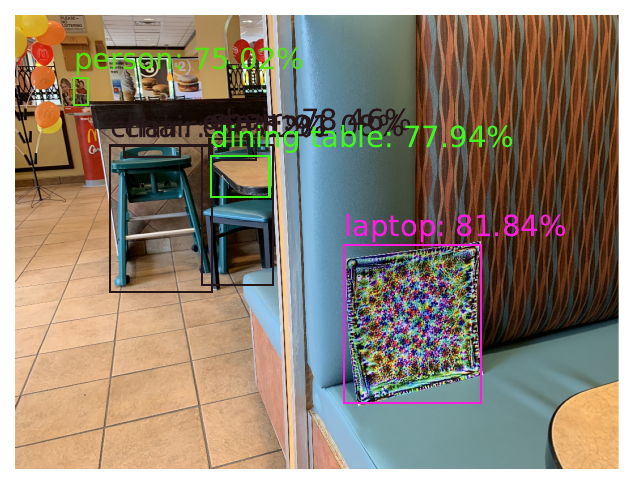}
        \end{subfigure}%
        \begin{subfigure}{0.5\columnwidth}
            \includegraphics[width=1.\textwidth]{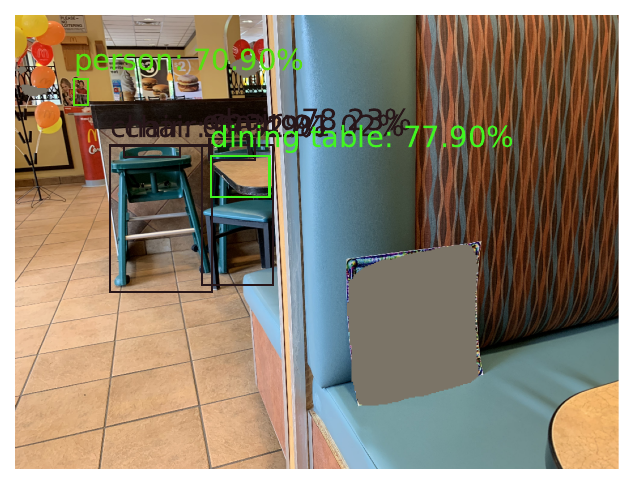}
        \end{subfigure}
        \caption{}
    \end{subfigure}
    \caption{(a) ROC analysis performed on the dataset including APRICOT images and 1000 COCO images. (b) The effect of \textit{Z-Mask} on an APRICOT image.}
    \label{fig:ROC_apricot}
\end{figure}

%% file: 5_conclusion.tex
\section{Conclusions}
\label{sec:conc}
This paper presented \textit{Z-Mask}, a method for masking and detecting physically-realizable adversarial examples.
This is accomplished by leveraging specific processing modules, such as the Spatial Pooling Refinement and the Fusion and Detection Block.
\textit{Z-Mask} is task-agnostic and was tested with OD and SS models, obtaining state-of-the-art results for both adversarial masking and detection on large datasets, as COCO and Cityscapes, and in real-world scenarios.
Furthermore, we strengthened the robustness of \textit{Z-Mask} by underlining the relation between over-activation and adversarial effect through an activation-aware patch optimization. 

As a future work, we plan to address an automatic selection of the shallow and deep layers involved in the over-activation analysis. 
Although the relation between over-activation and physical adversarial attacks is evident, it is less clear why certain model layers are more affected than others by this phenomenon. Addressing this task from a more theoretical perspective is not straightforward and requires further investigations.